\documentclass[11pt]{article}

\usepackage{acl}
\usepackage{microtype}
\usepackage{graphicx}
\usepackage{tikz}
\usetikzlibrary{arrows.meta,positioning,fit,backgrounds}
\usepackage{multirow}
\usepackage{booktabs}
\usepackage{tabularx}
\usepackage{amsmath,amssymb}
\usepackage{amsthm}
\usepackage{float}
\usepackage{dirtytalk}  
\usepackage{etoolbox}
\usepackage{hyperref}
\usepackage{comment}
\robustify{\say}
\usepackage[dvipsnames]{xcolor}

\usepackage{fontspec}
\newfontfamily\ethiopic{FreeSerif.ttf}[Path=./, Scale=1.0, AutoFakeBold=3]
\DeclareTextFontCommand{\textethiopic}{\ethiopic}
\newcommand{\gz}[1]{\textbf{\textethiopic{\csname gztext#1\endcsname}}}

\expandafter\def\csname gztextsbr_broke\endcsname{ሰበረ}
\expandafter\def\csname gztextsbr_frag\endcsname{ስብርባሪ}
\expandafter\def\csname gztextsbr_shatter\endcsname{ሰባበረ}
\expandafter\def\csname gztextsbr_passive\endcsname{ተሰበረ}
\expandafter\def\csname gztextsbr_instr\endcsname{መሰበሪያ}
\expandafter\def\csname gztextsbr_plural\endcsname{ተሰባበሩ}
\expandafter\def\csname gztextselam\endcsname{ሰላም}
\expandafter\def\csname gztextbasuma\endcsname{ባሱማ}
\expandafter\def\csname gztexthede\endcsname{ሄደ}
\expandafter\def\csname gztexthedu\endcsname{ሄዱ}
\expandafter\def\csname gztextbe\endcsname{በ}
\expandafter\def\csname gztextse\endcsname{ሰ}
\expandafter\def\csname gztextme\endcsname{መ}
\expandafter\def\csname gztextre\endcsname{ረ}
\expandafter\def\csname gztextte\endcsname{ተ}
\expandafter\def\csname gztextle\endcsname{ለ}
\expandafter\def\csname gztextye\endcsname{የ}
\expandafter\def\csname gztexthesyl\endcsname{ሄ}
\expandafter\def\csname gztextdusyl\endcsname{ዱ}
\expandafter\def\csname gztexthe\endcsname{ሀ}
\expandafter\def\csname gztextde\endcsname{ደ}
\expandafter\def\csname gztextha\endcsname{ሀ}
\expandafter\def\csname gztextsbrpl\endcsname{ሰባበሩ}
\expandafter\def\csname gztextsbpl\endcsname{ሰባበሩ}
\expandafter\def\csname gztextsbru\endcsname{ሰበሩ}
\expandafter\def\csname gztextsba\endcsname{ሰባ}
\expandafter\def\csname gztextsbat\endcsname{ሰባት}
\expandafter\def\csname gztextbru\endcsname{በሩ}
\expandafter\def\csname gztextsbe\endcsname{ሰበ}
\expandafter\def\csname gztextru\endcsname{ሩ}
\expandafter\def\csname gztextba\endcsname{ባ}
\expandafter\def\csname gztextsbbr\endcsname{ሰበበረ}
\expandafter\def\csname gztextsberu\endcsname{ሰበረ}
\expandafter\def\csname gztextsbdecg\endcsname{ሰበበረ}
\expandafter\def\csname gztextsbaa\endcsname{ሰባ}
\expandafter\def\csname gztextberu\endcsname{በሩ}
\expandafter\def\csname gztextsa\endcsname{ሰ}
\expandafter\def\csname gztextsu\endcsname{ሱ}
\expandafter\def\csname gztextsi\endcsname{ሲ}
\expandafter\def\csname gztextsae\endcsname{ሳ}
\expandafter\def\csname gztextsie\endcsname{ሴ}
\expandafter\def\csname gztexts\endcsname{ስ}
\expandafter\def\csname gztextso\endcsname{ሶ}
\expandafter\def\csname gztextba2\endcsname{ባ}
\expandafter\def\csname gztextbe2\endcsname{በ}
\expandafter\def\csname gztextru2\endcsname{ሩ}
\expandafter\def\csname gztextsbdec\endcsname{ሰበበረ}
\expandafter\def\csname gztextsbred\endcsname{ሰበረ}
\expandafter\def\csname gztextsbrA\endcsname{ሰበረ}
\expandafter\def\csname gztextsbrB\endcsname{ሰባበረ}
\expandafter\def\csname gztextsenA\endcsname{ልጁ መስታወቱን ሰበረ።}
\expandafter\def\csname gztextsenB\endcsname{ተማሪዎቹ ወንበሮቹን ሰባበሩ።}
\newcolumntype{L}[1]{>{\raggedright\arraybackslash}p{#1}}

\newcommand{\scheme}[1]{\texttt{\small\hyphenchar\font=`\_#1}}
\newcommand{\schemefns}[1]{\texttt{\scriptsize\hyphenchar\font=`\_#1}}
\newcommand{\splinter}{\textsc{Splinter}}

\newtheorem{definition}{Definition}
\newcommand{\Sys}{\textsc{Fidel}}
\newcommand{\Base}{\scheme{baseline}}
\newcommand{\SplOnly}{\scheme{splinter\_only}}
\newcommand{\Syl}{\scheme{syllabic}}
\newcommand{\SylSpl}{\scheme{syllabic\_then\_splinter}}
\newcommand{\CccSpl}{\scheme{cccvvv\_then\_splinter}}
\newcommand{\SplSyl}{\scheme{splinter\_then\_syllabic}}

\title{FIDEL: Abugida-Aware Pre-tokenization for Ethiopic Languages}

\author{Gebreslassie Teklu Reda \and Yuval Pinter \\
  Stein Faculty of Computer and Information Science \\
  Ben-Gurion University of the Negev \\
  Be’er Sheva, Israel\\
  \texttt{\{gebresla@post,uvp@cs\}.bgu.ac.il}}

\begin{document}
\maketitle

\begin{abstract}
Subword tokenizers assume that meaningful linguistic units can be recovered by splitting text at learned boundaries, an assumption better suited to concatenative morphology than to Semitic root-and-pattern systems. 
The challenge is acute for the Ethiopic languages, whose abugida script encodes consonant--vowel units as single characters and hides consonantal structure. 
We introduce \Sys{} (FIdäl DEcomposition and Linearization), a family of reversible abugida-aware pre-tokenization schemes that expose consonant and vowel-order structure, extend \splinter{} reductions from abjad scripts to the Ethiopic abugida, and evaluate them on three Ethiopic languages under controlled conditions, holding corpus, architecture, budget and protocol fixed so that tokenization is the only variable.
We report intrinsic tokenizer statistics for BPE and UnigramLM across vocabulary sizes alongside downstream NLP tasks, including transfer scenarios.
We find that \Sys{} reduction improves news classification consistently and named-entity recognition at larger capacity, while sentiment analysis performance declines under the more aggressive schemes, and extractive question answering performance is preserved.
\end{abstract}

\section{Introduction}
\label{sec:intro}

Every language model begins its operation with tokenization, the mapping from character sequences to integer sequences that
determine what the model can learn about word structure. The dominant algorithms~\citep{schuster2012,sennrich-etal-2016-neural,kudo-2018-subword} share a concatenative assumption: an ID only ever represents runs of adjacent characters, so the units a tokenizer can discover are contiguous substrings. 
In cases where morphology is expressed by interleaving rather than concatenation, no concatenative schema can recover morphemes.
Such is the case with Semitic languages, which build words from a consonantal root combined with a mostly vocalic template.
In an abjad script, such as that representing Hebrew or Arabic, vowels are largely unwritten, so the written form
approximates the skeleton directly, and recent work exploits this to improve tokenization~\citep{gazit-etal-2025-splintering}.

\begin{table}[t]
\centering\small
\begin{tabular}{@{}l l@{}}
\toprule
\multicolumn{2}{@{}l}{(a) \; \gz{senA}} \\
\multicolumn{2}{@{}l}{\hspace{1.4em}\say{The boy broke the mirror.}} \\[2pt]
\multicolumn{2}{@{}l}{(b) \; \gz{senB}} \\
\multicolumn{2}{@{}l}{\hspace{1.4em}\say{The students broke the chairs into pieces.}} \\
\midrule
\textbf{Script} & \textbf{Challenge and solution} \\
Alphabet & Skeleton visible; with concatenative \\
         & morphology, subwords generally suffice. \\
Abjad    & Skeleton visible but intermixed with \\
         & morphology; exposed by \splinter{}. \\
Abugida  & Skeleton fused into consonant--vowel \\
         & codepoints; exposed by \Sys{}, after \\
         & which \splinter{} applies. \\
\bottomrule
\end{tabular}
\caption{Examples of Amharic text exemplifying its nonconcatenative morphology.
The root \gz{sberu} \say{break} appears in both sentences, but in (a) it comprises a contiguous segment whereas in (b) it is broken by the templatic \gz{ba2} and by the vowel order of its final consonant (\gz{re}~$\rightarrow$~\gz{ru2}).
The lower block compares abugida with related script types.}
\label{tab:sbr-family}
\end{table}

\begin{figure*}[t]
\centering
\small
\begin{tikzpicture}[
  scale=0.77, transform shape,
  cell/.style={draw, minimum height=7mm, minimum width=6.5mm, inner xsep=2.5pt,
               fill=gray!12, font=\small},
  grp/.style={draw, inner sep=1.6pt},
  lbl/.style={font=\small},
  sub/.style={font=\small},
  ar/.style={-{Latex[length=2.2mm]}, very thick}]

\node[font=\bfseries, anchor=west] at (-0.3,2.6) {Standard Flow:};
\node[lbl] at (0.35,1.75) {Input};
\draw[ar] (-0.3,1.1) -- (0.75,1.1);
\node[cell] (a1) at (1.35,1.1) {\gz{sa}};
\node[cell, right=0.5mm of a1] (a2) {\gz{ba2}};
\node[cell, right=0.5mm of a2] (a3) {\gz{be2}};
\node[cell, right=0.5mm of a3] (a4) {\gz{ru2}};
\begin{scope}[on background layer]\node[grp, fit=(a1)(a2)(a3)(a4)] (A) {};\end{scope}
\node[sub, below=1.2mm of A] {Original text};
\node[lbl] at (5.2,1.75) {Tokenizer};
\draw[ar] (A.east) -- (6.3,1.1);
\node[cell, minimum width=12mm] (b1) at (7.2,1.1) {\gz{sbaa}};
\node[cell, minimum width=12mm, right=0.5mm of b1] (b2) {\gz{beru}};
\begin{scope}[on background layer]\node[grp, fit=(b1)(b2)] (B) {};\end{scope}
\node[sub, below=1.2mm of B] {Segmented text};
\node[lbl] at (9.9,1.75) {Embedding};
\draw[ar] (B.east) -- (11.0,1.1);
\node[draw, minimum height=8mm, minimum width=9mm, font=\bfseries] at (11.6,1.1) {LM};

\node[font=\bfseries, anchor=west] at (-0.3,-0.5) {\Sys{} Flow:};
\node[lbl] at (0.35,-1.35) {Input};
\draw[ar] (-0.3,-2.0) -- (0.95,-2.0);
\node[cell] (c1) at (1.35,-2.0) {\gz{sa}};
\node[cell, right=0.5mm of c1] (c2) {\gz{ba2}};
\node[cell, right=0.5mm of c2] (c3) {\gz{be2}};
\node[cell, right=0.5mm of c3] (c4) {\gz{ru2}};
\begin{scope}[on background layer]\node[grp, fit=(c1)(c2)(c3)(c4)] (C) {};\end{scope}
\node[sub, below=1.2mm of C] {Original text};

\node[lbl] at (4.0,-1.35) {Decompose};
\draw[ar] (C.east) -- (5.2,-2.0);
\node[cell] (d1) at (5.6,-2.0) {\gz{sa}};
\node[cell, right=0.5mm of d1] (d2) {\gz{be2}};
\node[cell, right=0.5mm of d2] (d3) {$V_3$};
\node[cell, right=0.5mm of d3] (d4) {\gz{be2}};
\node[cell, right=0.5mm of d4] (d5) {\gz{re}};
\node[cell, right=0.5mm of d5] (d6) {$V_1$};
\begin{scope}[on background layer]\node[grp, fit=(d1)(d2)(d3)(d4)(d5)(d6)] (D) {};\end{scope}
\node[sub, below=1.2mm of D] {Decomposed text};

\node[lbl] at (9.5,-1.35) {Reduce};
\draw[ar] (D.east) -- (10.5,-2.0);
\node[cell] (e1) at (10.9,-2.0) {\gz{sa}};
\node[cell, right=0.5mm of e1] (e2) {\gz{be2}};
\node[cell, right=0.5mm of e2] (e3) {\gz{re}};
\node[cell, right=0.5mm of e3] (e4) {$R_1$};
\node[cell, right=0.5mm of e4] (e5) {$R_2$};
\begin{scope}[on background layer]\node[grp, fit=(e1)(e2)(e3)(e4)(e5)] (E) {};\end{scope}
\node[sub, below=1.2mm of E] {Reduced text};

\node[lbl] at (14.1,-1.35) {Tokenizer};
\draw[ar] (E.east) -- (14.7,-2.0);
\node[cell, minimum width=12mm] (f1) at (15.4,-2.0) {\gz{sberu}};
\node[cell, right=0.5mm of f1] (f2) {$R_1R_2$};
\begin{scope}[on background layer]\node[grp, fit=(f1)(f2)] (F) {};\end{scope}
\node[sub, below=1.2mm of F] {Segmented text};
\draw[ar] (F.east) -- (17.75,-2.0);
\node[draw, minimum height=8mm, minimum width=9mm, font=\bfseries] at (18.2,-2.0) {LM};
\node[lbl] at (17.1,-1.35) {Embedding};
\end{tikzpicture}
\caption{An Amharic language model pipeline for \gz{sbpl} (\emph{s\"ababb\"aru},
\say{they shattered}): the standard flow for a 16k-vocabulary BPE tokenizer, with and without incorporating \Sys{}.
The standard flow yields pieces that share nothing with the segmentation of the related
\gz{sbr_broke}, so the root is never exposed; \Sys{} decomposes and reduces the word to that
root, which the tokenizer then emits as a single shared piece.}
\label{fig:pipeline}
\end{figure*}
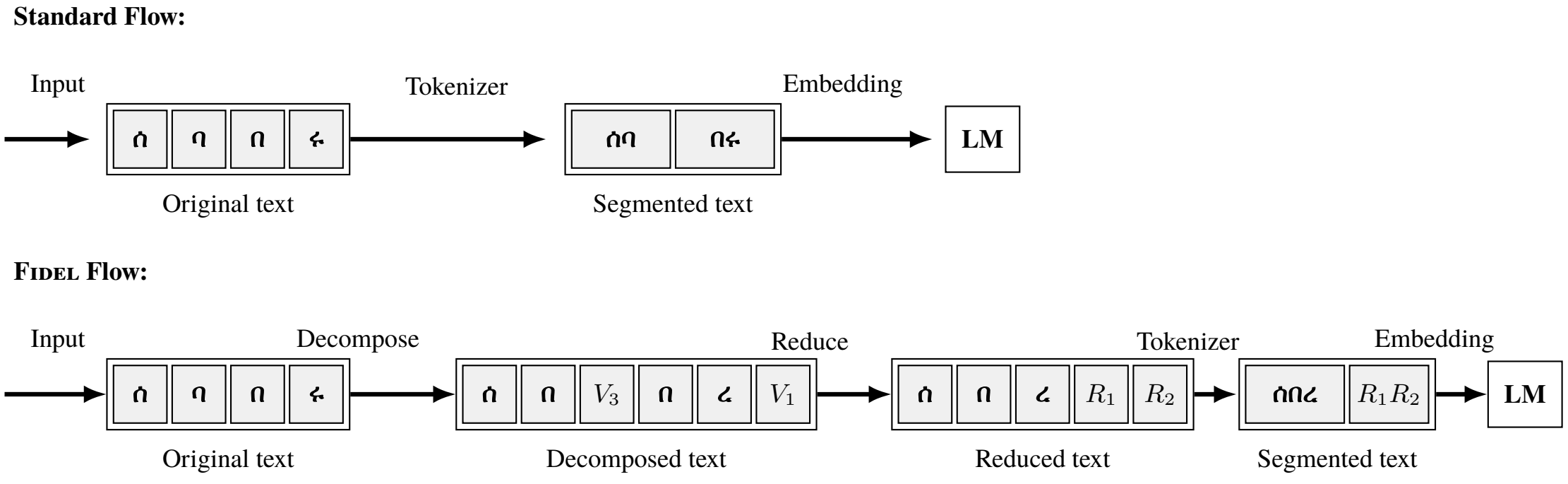

The Ethiopic languages present a harder case.
Amharic, Tigrinya and Ge'ez are written in the \emph{fid\"al}, an abugida of 33 base consonants each realizable in seven vowel orders.
In Unicode, each character is represented as one indivisible codepoint fusing a consonant with a vowel: \gz{be} is not \say{b} followed by \say{e}, but a single symbol denoting the combined form.
In the example in \autoref{tab:sbr-family}, both verbs meaning \say{broke} derive from the root \gz{se}--\gz{be}--\gz{re}, but in \gz{sbpl} the frequentative template inserts \gz{ba2} and raises the vowel order of the final consonant~(\gz{re}~$\rightarrow$~\gz{ru2}).
A tokenizer can only be exposed to the syllabic characters of each form and never the common consonantal sequence, because that skeleton is not a subsequence of the written string.

This is also why \splinter{}~\citep{gazit-etal-2025-splintering}, which isolates roots in abjads by deleting templatic material, cannot be applied directly to Ethiopic scripts.
Consequently, morphologically related words may be segmented into unrelated subword units, limiting the model's ability to generalize across forms derived from the same root.
Existing Ethiopic resources treat tokenization as fixed infrastructure~\citep{gezmu-etal-2022-extended,belay2022,tonja-etal-2024-ethiollm}, so no study isolates its effect for these languages. 
We therefore introduce \Sys{}, a reversible pre-tokenization framework for the Ethiopic abugida that decomposes each fid\"al character into a canonical consonant and an explicit vowel-order marker, exposing the hidden consonantal structure and enabling \splinter{}-style reductions to recover shared root patterns that are inaccessible while consonants and vowels remain fused.
As shown in \autoref{fig:pipeline}, the resulting representation can be processed by any ordinary subword tokenizer and decoded back into standard fid\"al thanks to systematic recording of the vowel-order decomposition and reduction operations.
We evaluate several ways of combining the two operations, varying both representation and operation order, and find that the resulting representations improve token efficiency at larger vocabularies and benefit news classification and named-entity recognition downstream.


Our contributions are: \Sys{}, a reversible pre-tokenization framework for abugida scripts with a formal decoding procedure; an adaptation of \splinter{} from abjads to an abugida; a controlled comparison across three languages, two model scales, two vocabulary sizes and four task families; and intrinsic tokenizer analyses for two subword algorithms across a range of vocabulary sizes, reported alongside the downstream results they are often used to predict.


\section{Background and Related Work}
\label{sec:background}

\paragraph{Subword tokenization.}
The three main tokenization algorithms:
byte-pair encoding~\citep[BPE;][]{sennrich-etal-2016-neural} which builds a vocabulary by iteratively merging frequent adjacent token pairs;
WordPiece~\citep{schuster2012} which merges by likelihood gain;
and UnigramLM~\citep{kudo-2018-subword} which prunes a large candidate vocabulary,
all share a concatenative assumption.  
Only adjacent symbols form tokens, either through merges or by initial vocabulary construction.
This makes them less effective at capturing morphological structures in which related elements are interleaved rather than concatenated.
In addition, standard tokenizers may violate script-specific orthographic constraints in abugida writing systems, producing invalid token boundaries~\citep{kapila2026brahmic}.
Tokenizer quality affects downstream performance~\citep{rust-etal-2021-good}, and subwords align poorly with morphemes in morphologically complex languages~\citep{klein2020}.

\paragraph{The fid\"al.} The Ethiopic (Ge'ez) script is an abugida in which each fid\"al character encodes a consonant--vowel sequence rather than a bare phoneme~\citep{daniels1996world}. 
The consonantal skeleton of Ge'ez admits seven vowel orders, realized as modifications of a consonant's base shape~\citep{dillmann1974ethiopic,leslau1987comparative}.
Thus, consonant and vowel information are fused within a single character. 
The consonantal skeleton is therefore not directly represented in the written string; the verbs \gz{hede}~(\say{he went}) and \gz{hedu}~(\say{they went}) differ only in the vowel order of the final character but share no symbol in that position.
This fusion of consonants and vowels prevents structure from being directly accessible to a tokenizer.

\paragraph{\splinter{}.} \citet{gazit-etal-2025-splintering} address nonconcatenative morphology through a reversible input transformation that relinearizes text by iteratively pruning characters from words and re-appending them at the end. 
The learned pruning operations specify both a character and its location in the word, and the transformed text is processed by an unmodified subword tokenizer. 
The authors report improvements on downstream tasks for Hebrew and Arabic, both of which exhibit nonconcatenative root-and-pattern morphology.
In these languages, roots are represented by separable consonantal characters, while vowels may be unwritten or represented separately~\citep{daniels1996world}, allowing \splinter{}'s character-deletion operations to expose the roots.
However, \splinter{}'s character-deletion operations cannot be applied directly to Ethiopic abugida, where consonant and vowel information are fused within a single fid\"al character.
Deleting a whole character would therefore remove both consonant and vowel information, rather than isolating the consonantal root. 
Extending \splinter{} to Ethiopic therefore requires a prior decomposition step that transforms each fid\"al into separable consonant and vowel elements before reduction can apply.

\paragraph{Ethiopic script representations and Resources.}
Representing an abugida as consonant--vowel pairs is long established outside tokenization.
The SERA scheme~\citep{firdyiwek1997sera} maps each fid\"al character to an ASCII digraph for interchange, and finite-state morphological analysers operate over decomposed representations~\citep{gasser-2009-semitic,gasser2011hornmorpho}. 
Beyond representation and morphological analysis, computational treatments of Ethiopic languages include stemming~\citep{alemu-argaw-asker-2007-amharic} and corpus construction~\citep{gezmu-etal-2018-contemporary}, while EthioLLM~\citep{tonja-etal-2024-ethiollm} provides encoders with a 256k subword vocabulary and benchmarks, including MasakhaNER~\citep{adelani-etal-2021-masakhaner} and AfriSenti~\citep{muhammad-etal-2023-afrisenti}.

\paragraph{Tokenization for Ge'ez-script languages.}
Closest to our work is MoVoC~\citep{teklehaymanot-etal-2025-movoc}, which builds a morpheme-aware subword vocabulary for Ge'ez-script languages by combining supervised morphological analysis with BPE over annotated morpheme data.
It modifies the \emph{vocabulary} and depends on annotation, whereas we modify the \emph{input} and require none, leaving the subword algorithm unmodified, so that differences are attributable to the representation rather than to the segmenter. 
They find that morphology-aware segmentation improves intrinsic measures without yielding significant translation gains, consistent with our observation that intrinsic statistics do not straightforwardly predict downstream behavior.


\section{Method}
\label{sec:method}

Every \Sys{} scheme operates as a \emph{pre-tokenization} step: raw text is transformed and passed
to a tokenizer built using SentencePiece~\citep{kudo-richardson-2018-sentencepiece}.
We compare six pre-tokenization schemes that combine text representations and \splinter{}-based reduction in different ways.
We describe the text representations in Section~\ref{sec:representations}, the reduction applied in Section~\ref{sec:splinter}, and the reversible decoding procedure in Section~\ref{sec:reversibility}.
\autoref{tab:schemes} presents the conditions we compare and shows each applied to the same input.

\subsection{Consonant--Vowel Representations}
\label{sec:representations}

\begin{table*}[t]
\centering\small
\begin{tabularx}{\textwidth}{@{}l c l c l X@{}}
\toprule
\textbf{Scheme} & & \textbf{Decomposition} & & \textbf{Reduction} & \textbf{Definition}\\
\midrule
\Base{} & & --- & & ---
  & Raw fid\"al, no transformation: \gz{sbpl}.\\[2pt]
\SplOnly{} & & --- & & \gz{sbru}\;$R$
  & Reduction only, applied directly to fid\"al.\\[2pt]
\Syl{} & & \gz{se}\gz{be}$V_3$\gz{be}\gz{re}$V_1$ & & ---
  & Decomposition only, no reduction.\\
\midrule
\SylSpl{}
  & & \gz{se}\gz{be}$V_3$\gz{be}\gz{re}$V_1$
  & $\longrightarrow$
  & \gz{sberu}\;$R_1R_2$
  & Interleaved (CVCV) decomposition, then reduction.\\[2pt]
\CccSpl{}
  & & \gz{sbdecg}\;$V_{3,2}V_{1,4}$
  & $\longrightarrow$
  & \gz{sberu}\;$R_1R_2$
  & Grouped (CCCVVV) decomposition, then reduction.\\[2pt]
\SplSyl{}
  & & \gz{sberu}$V_1$\;$R$
  & $\longleftarrow$
  & \gz{sbru}\;$R$
  & Reduction first, then decomposition (operation-order ablation).\\
\bottomrule
\end{tabularx}
\caption{The pre-tokenization schemes, each applied to \gz{sbpl}
(\emph{s\"ababb\"aru}, \say{they broke into pieces }).
The arrow gives the order in which the two operations are applied, and a dash marks an operation the scheme does not perform.
$R_1$ and $R_2$ are reduction markers, each a single codepoint recording one operation. Appendix~\ref{app:pipelines} presents the same conditions as pipelines.}
\label{tab:schemes}
\end{table*}
The fid\"al is decomposed into a base consonant and a separate vowel marker, giving two elements per syllable. The \emph{interleaved} form (CVCV) places each vowel marker directly after its consonant, preserving syllable boundaries; the \emph{grouped} form (CCCVVV) emits all consonants first and all vowel markers afterwards:
\begin{center}
\begin{tabular}{cc}
\gz{sbpl} $\rightarrow$ \gz{se}\; \gz{be}$V_3$\; \gz{be}\; \gz{re}$V_1$ & (interleaved)\\[2pt]
\gz{sbpl} $\rightarrow$ \gz{se}\gz{be}\gz{be}\gz{re}\,$V_{3,2}V_{1,4}$ & (grouped)
\end{tabular}
\end{center}
In the interleaved representation, each marker follows its consonant, so adjacency determines the pairing. In the grouped representation, markers follow all consonants and explicitly identify their positions. In \(V_{o,i}\), \(o\) denotes the vowel order and \(i\) the one-based consonant position. Thus, \(V_{3,2}V_{1,4}\) assigns vowel order 3 to the second consonant and vowel order 1 to the fourth; unmarked consonants retain the first order.
In both, the skeleton \gz{se}, \gz{be}, \gz{re} becomes explicit, and the frequentative template and plural inflection that distinguish \gz{sbpl} from \gz{sbr_broke} are isolated in separate markers rather than buried in a distinct codepoint. 
We implement markers using unused CJK codepoints, following \citet{gazit-etal-2025-splintering}.
\autoref{app:sbr} presents all \emph{s-b-r} forms under this decomposition.

\paragraph{First-order canonicalization.} The consonantal component is always written in the canonical \emph{first-order} fid\"al form, with the vowel order carried separately.
A \emph{vowel order} is the position of a character within the seven-member series that a base consonant forms: \gz{se} is the first order of the consonant \emph{s}.
The remaining six orders realize the same consonant with each of the other six vowels, giving the seven-member series \gz{se}, \gz{su}, \gz{si}, \gz{sae}, \gz{sie}, \gz{s}, \gz{so}.
Vowel orders are indexed from zero, $V_0$ through $V_6$; $V_0$ is canonical, meaning that a consonant already in first order carries no marker. 
Thus, as shown in the decomposition column of \autoref{tab:schemes}, the first and third consonants are already in first order and therefore written bare.
The absence of a marker is itself informative, which is what makes the decoding rule (see Section~\ref{sec:reversibility}) well-defined: an unmarked consonant is reconstructed in first order on both sides of the mapping.
 
\subsection{Adapting \splinter{} to an Abugida}
\label{sec:splinter}

In our main setup, \splinter{} is applied over decomposed representations rather than over raw fid\"al.
\autoref{tab:schemes} presents the \SplOnly{} scheme, which reduces \gz{sbpl} only to \gz{sbru}, whereas both composed schemes reach the root \gz{sberu}.
Reaching the root requires manipulating vowel-order information (\gz{ru} $\rightarrow$ \gz{re}), which is impossible while consonants and vowels remain fused.
Only after decomposition separates consonants from vowel orders can \splinter{} remove templatic material and recover the root.

The transformation in \Sys{} proceeds in two stages.
Decomposition rewrites the word over the  consonant alphabet, giving \gz{se}\,\gz{be}$V_3$\,\gz{be}\,\gz{re}$V_1$. 
Reduction then operates on this sequence. 
The frequentative element \gz{be}$V_3$ is removed and the final vowel marker $V_1$ is lowered to first order, yielding the root \gz{sberu} (\emph{s\"abb\"ar\"a}, \say{he broke}), which is substantially more frequent in the corpus. Each operation is recorded as a reduction marker containing the removed material and its position, so that $R_1$ restores \gz{be}$V_3$ and $R_2$ restores $V_1$. 
The encoded form therefore consists of the consonantal skeleton together with the reduction markers; decoding applies the recorded operations in reverse order and reconstructs \gz{sbpl} exactly.
In the grouped form, markers follow all bases: \gz{sbdecg}$V_{3,2}V_{1,4}$ undergoes the same reductions, yielding \gz{sberu}$R_1R_2$.
Decoding is likewise the same replay in reverse: $R_2$ returns order 1 to the third consonant, and $R_1$ reinserts \gz{be} with order 3 as the second, recomposing \gz{sbpl}.
Both representations use the same reduction and reconstruction procedure; they differ only in how vowel markers are associated with consonants: by adjacency in the interleaved form and by explicit position in the grouped form.

Reductions are applied iteratively, allowing words with multiple affixes to undergo several reduction steps. 
The process halts when no more frequent attested intermediate form exists, preventing spurious root recovery.

Two constraints keep reductions morphological. First, a candidate is accepted only if the result is itself an attested word that is more frequent than the input, following \citet{gazit-etal-2025-splintering}. 
Second, the result must retain at least three units, preventing short words from collapsing into single frequent characters that are orthographic rather than morphological relatives.
Reduction statistics are computed per language; otherwise, a form frequent in one language may incorrectly attract reductions from another.
For example, the Tigrinya word \gz{sbat} (``people'', frequency 23) would reduce to the Amharic \gz{sba} (``seventy'',
frequency 54) under pooled statistics, even though the two are unrelated.

The decomposition schemes differ in what they can expose. \Syl{} separates vowel-order information from the consonants, increasing overlap between related forms but leaving internal templatic material intact.
\SplOnly{} applies reduction while vowels remain fused, so it may halt at an intermediate form.
Only the two schemes that both decompose and reduce, \SylSpl{} and \CccSpl{}, reach the root, because decomposition makes the consonants explicit before reduction removes templatic material.

\subsection{Reversibility and decoding}
\label{sec:reversibility}

Because the representation separates consonant identities from vowel information, a deterministic decoding procedure is required to reconstruct valid Ethiopic words. 
Encoding is injective, and decoding its output is a deterministic inverse: each consonant takes the vowel order specified by its marker, unmarked consonants remain in first order, and reduction markers are undone in reverse order.
We require of every scheme that
\begin{equation*}
\mathrm{Decode}(\mathrm{Encode}(W)) = W
\end{equation*}
\noindent for every valid Ethiopic abugida word $W$, so that preprocessing loses no information and predictions over the transformed stream can be scored against the original surface form.
When a language model generates malformed sequences, such as markers referring to nonexistent consonant positions or missing vowel instructions, the decoder uses the projection procedure described in  \autoref{app:decoding} to handle these cases.

\paragraph{Verification.} Reversibility is checked by round-trip reconstruction rather than assumed.
On the Amharic evaluation split, \scheme{cccvvv\_then\_splinter} reconstructed 652{,}966 of 652{,}966 whitespace-delimited tokens exactly. 
Three classes of edge case are handled explicitly: Ethiopic punctuation (U+1360--U+1368), trimmed before encoding and reattached afterwards; foreign-script material whose codepoints overlap the CJK marker blocks, passed through untouched; and labiovelars, which fall outside the regular seven-order grid. For any form the encoder cannot verify, it returns the input unchanged, so identity holds by construction.

\section{Intrinsic Tokenizer Analysis}
\label{sec:intrinsic}

We first evaluate each representation independently of any trained model. 
We report five measures, each calculated over a corpus: \textbf{R\'enyi efficiency}~\citep{zouhar-etal-2023-tokenization} penalizes token distributions skewed towards very high- or very low-frequency tokens (higher is better).\footnote{Note that \citet{cognetta-etal-2024-two} show this metric can be raised adversarially while downstream performance degrades.}
\textbf{Tokens per word}, also called subword fertility~\citep{rust-etal-2021-good}, is the
mean number of tokens a whitespace-delimited word is split into;
\textbf{4+ token words} is the proportion of words segmented into four or more tokens; and
\textbf{1-char tokens} the proportion of tokens consisting of a single character. 
All three are indicators of compression efficiency (lower is better).
Finally,
\textbf{distinct neighbors} measures contextual coherence~\citep{yehezkel-pinter-2023-incorporating}, the number of distinct tokens each token co-occurs with within a distance of two, averaged over the vocabulary.
Fewer contexts make a token's environments more distinguishable, so lower is better.
Together, these measures capture different and partly conflicting aspects of tokenizer quality.

\begin{table*}[!t]
\centering\small
\begin{tabular}{@{}lrrrrr@{}}
\toprule
\textbf{Scheme} & \textbf{\begin{tabular}[c]{@{}r@{}}R\'enyi\\ efficiency\end{tabular}} & \textbf{\begin{tabular}[c]{@{}r@{}}Tokens\\ per word\end{tabular}} & \textbf{\begin{tabular}[c]{@{}r@{}}4+ token\\ words\end{tabular}} & \textbf{\begin{tabular}[c]{@{}r@{}}1-char\\ tokens\end{tabular}} & \textbf{\begin{tabular}[c]{@{}r@{}}Distinct\\ Neighbors\end{tabular}} \\
\midrule
\multicolumn{6}{@{}l}{\emph{BPE}}\\
\scheme{baseline}                  & 0.654 & 1.353 & \textbf{1.98\%} & 1.89\% & 7286\\
\scheme{syllabic\_then\_splinter} & 0.643 & \textbf{1.323} & 2.13\% & \textbf{1.19}\% & 6329\\
\scheme{splinter\_then\_syllabic} & 0.654 & 1.362 & 2.33\% & 1.21\% & 7315\\
\scheme{splinter\_only}            & 0.655 & 1.367 & 2.41\% & 1.95\% & 7304\\
\scheme{cccvvv\_then\_splinter}    & \textbf{0.668} & 1.498 & 5.34\% & 2.40\% & 8421\\
\scheme{syllabic}                  & 0.571 & 2.391 & 5.12\% & 1.60\% & \textbf{5628}\\
\midrule
\multicolumn{6}{@{}l}{\emph{UnigramLM}}\\
\scheme{baseline}                  & 0.547 & 1.382 & 2.67\% & 2.53\% & 6988\\
\scheme{syllabic\_then\_splinter} & 0.574 & \textbf{1.346} & \textbf{2.49\%} & \textbf{1.32\%} & 5865\\
\scheme{splinter\_then\_syllabic} & \textbf{0.600} & 1.377 & 2.59\% & 1.43\% & 7104\\
\scheme{splinter\_only}            & 0.598 & 1.382 & 2.81\% & 2.43\% & 7082\\
\scheme{cccvvv\_then\_splinter}    & 0.596 & 1.503 & 6.17\% & 3.14\% & 7572\\
\scheme{syllabic}                  & 0.548 & 2.422 & 5.36\% & 1.69\% & \textbf{5766}\\
\bottomrule
\end{tabular}
\caption{Intrinsic statistics at 64k vocabulary for both subword algorithms. Bold marks the
best value per column within each block; lower is better
for all measures except R\'enyi efficiency.}
\label{tab:intrinsic64}
\end{table*}

\begin{table*}[!t]
\centering\small
\begin{tabular}{@{}lrrrrr@{}}
\toprule
\textbf{Scheme} & \textbf{\begin{tabular}[c]{@{}r@{}}R\'enyi\\ efficiency\end{tabular}} & \textbf{\begin{tabular}[c]{@{}r@{}}Tokens\\ per word\end{tabular}} & \textbf{\begin{tabular}[c]{@{}r@{}}4+ token\\ words\end{tabular}} & \textbf{\begin{tabular}[c]{@{}r@{}}1-char\\ tokens\end{tabular}} & \textbf{\begin{tabular}[c]{@{}r@{}}Distinct\\ Neighbors\end{tabular}} \\
\midrule
\multicolumn{6}{@{}l}{\emph{64k vocabulary}}\\
\scheme{baseline}                  & 0.654 & 1.353 & \textbf{1.98\%} & 1.89\% & 7286\\
\scheme{syllabic\_then\_splinter} & 0.643 & \textbf{1.323} & 2.13\% & \textbf{1.19}\% & 6329\\
\scheme{splinter\_then\_syllabic} & 0.654 & 1.362 & 2.33\% & 1.21\% & 7315\\
\scheme{splinter\_only}            & 0.655 & 1.367 & 2.41\% & 1.95\% & 7304\\
\scheme{cccvvv\_then\_splinter}    & \textbf{0.668} & 1.498 & 5.34\% & 2.40\% & 8421\\
\scheme{syllabic}                  & 0.571 & 2.391 & 5.12\% & 1.60\% & \textbf{5628}\\
\midrule
\multicolumn{6}{@{}l}{\emph{32k vocabulary}}\\
\scheme{baseline}                  & 0.688 & 1.495 & \textbf{3.33\%} & 2.84\% & 6480\\
\scheme{syllabic\_then\_splinter} & 0.675 & \textbf{1.471} & 3.89\% & \textbf{1.74}\% & 5980\\
\scheme{splinter\_then\_syllabic} & 0.688 & 1.512 & 4.09\% & 1.77\% & 6464\\
\scheme{splinter\_only}            & 0.688 & 1.519 & 4.22\% & 2.93\% & 6439\\
\scheme{cccvvv\_then\_splinter}    & \textbf{0.711} & 1.725 & 9.57\% & 3.37\% & 7319\\
\scheme{syllabic}                  & 0.582 & 2.496 & 7.82\% & 1.98\% & \textbf{4713}\\
\midrule
\multicolumn{6}{@{}l}{\emph{16k vocabulary}}\\
\scheme{baseline}                  & 0.722 & \textbf{1.694} & \textbf{5.68\%} & 4.18\% & 5207\\
\scheme{syllabic\_then\_splinter} & 0.708 & 1.696 & 7.13\% & 2.68\% & 4900\\
\scheme{splinter\_then\_syllabic} & 0.720 & 1.731 & 7.29\% & 2.63\% & 5069\\
\scheme{splinter\_only}            & 0.720 & 1.744 & 7.48\% & 4.51\% & 5027\\
\scheme{cccvvv\_then\_splinter}    & \textbf{0.749} & 2.040 & 15.84\% & 4.79\% & 5686\\
\scheme{syllabic}                  & 0.590 & 2.630 & 12.04\% & \textbf{2.69\%} & \textbf{3562}\\
\midrule
\multicolumn{6}{@{}l}{\emph{8k vocabulary}}\\
\scheme{baseline}                  & 0.745 & \textbf{1.981} & \textbf{10.15\%} & 6.75\% & 3655\\
\scheme{splinter\_then\_syllabic} & 0.745 & 2.086 & 13.73\% & 4.76\% & 3351\\
\scheme{syllabic\_then\_splinter} & 0.735 & 2.090 & 14.32\% & 4.31\% & 3393\\
\scheme{splinter\_only}            & 0.742 & 2.104 & 14.07\% & 7.60\% & 3322\\
\scheme{cccvvv\_then\_splinter}    & \textbf{0.773} & 2.496 & 25.84\% & 6.82\% & 3877\\
\scheme{syllabic}                  & 0.590 & 2.810 & 18.04\% & \textbf{4.14\%} & \textbf{2410}\\
\bottomrule
\end{tabular}
\caption{Intrinsic statistics for BPE across vocabulary sizes, following the protocol of
\citet{gazit-etal-2025-splintering}. Bold marks the best value per column within each block; lower is better
for all measures except R\'enyi efficiency.}
\label{tab:intrinsic}
\end{table*}

\autoref{tab:intrinsic64} summarises both BPE and UnigramLM at 64k vocabulary, and
\autoref{tab:intrinsic} focuses on vocabulary size effects in BPE; UnigramLM numbers are reported in
\autoref{app:intrinsic}. 
The interleaved scheme \scheme{syllabic\_then\_splinter} is the most token-efficient at the
larger vocabularies, slightly ahead of the untransformed baseline, for both BPE and UnigramLM.
At 16k and below, the baseline regains the advantage, so any benefit of decomposition depends on there being vocabulary capacity to absorb the additional symbols. 
\scheme{syllabic}, which marks vowel order without applying reduction, is the least token-efficient scheme at every size, producing roughly 1.8 times as many tokens per word as the baseline; because its markers encode position as well as vowel order, its marker inventory is large and each marker is individually rare.
It nevertheless yields the fewest distinct neighbors at every size, so the two measures point in opposite directions for this scheme.
More generally, no single measure orders the schemes as the others do, which is consistent with the caution of \citet{schmidt-etal-2024-tokenization} against assessing tokenizers by compression alone.

\section{Downstream Setup}
\label{sec:setup}

We further evaluate the \Sys{} pre-tokenization schemes by training language models from scratch on each representation and fine-tuning them on downstream tasks.
Holding the corpus, architecture, optimization budget and evaluation protocol fixed leaves the tokenization as the only variable, so any difference in downstream performance is attributable to the representation.

\paragraph{Corpus and models.} 
We pretrain encoders with pre-layer-normalization from
scratch using masked language modeling~\citep{devlin-etal-2019-bert} on a 385\,MB Ge'ez-script corpus
(662 files, 1{,}447{,}588 lines), identical across conditions.
Two scales are used,
\emph{small} ($\sim$91M parameters) and \emph{large} ($\sim$165M), each with vocabularies of 16k and 64k.
We optimize using AdamW~\citep{loshchilov2017fixing} with warmup and cosine decay. 
Downstream fine-tuning updates the full encoder under an identical budget, and the best-validation checkpoint is evaluated. Further details are provided in \autoref{app:training-hyperparameters}.
Experiments were conducted on an HPC cluster using NVIDIA RTX~3090 GPUs (24\,GB), one GPU per run.\footnote{Pretraining the 24 encoder configurations (six schemes $\times$ two scales $\times$ two vocabulary sizes) took 2.0 to 5.6 days per run and roughly 90 GPU-days in total, though jobs ran concurrently across available GPUs, so wall-clock time was substantially shorter.}
\paragraph{Tasks.} Evaluation covers named-entity
recognition~\citep[EthioNER, MasakhaNER;][]{tonja-etal-2024-ethiollm,adelani-etal-2021-masakhaner},
sentiment analysis~\citep[AfriSenti;][]{muhammad-etal-2023-afrisenti}, news
classification~\citep[MasakhaNEWS;][]{adelani-etal-2023-masakhanews} and
part-of-speech tagging~\citep[UD\_Amharic-ATT;][]{seyoum-etal-2018-universal}.
Named-entity recognition is reported as \textbf{entity-F1}, and the remaining tasks as
weighted F1. 
finding no entities would still exceed 0.90 weighted-F1.


\paragraph{Controlled comparison.} All causal claims are made against our internal baseline,
in which corpus, architecture, budget and protocol are identical and only tokenization
differs.
Published systems differ in data, architecture and training regime simultaneously, so no difference against them could be attributed to tokenization; their figures are cited as
context only.
Reported values are medians over the runs available for each configuration.

\section{Downstream Results}
\label{sec:results}

We report two configurations of the large model. The \textbf{64k} vocabulary gives the subword algorithm enough capacity to absorb the additional symbols that decomposition introduces, so marker-bearing schemes are not penalized for inventory size; this is the setting in which the intrinsic statistics of Section~\ref{sec:intrinsic} favor the transformed representations.
The \textbf{16k} vocabulary is the constrained setting, where every marker competes with a stem for vocabulary slots, and where the baseline regains its intrinsic advantage.
Both are realistic, corresponding roughly to monolingual model budgets and script-within-multilingual model vocabulary budgets, respectively.
Comparing the two isolates how much of any downstream effect depends on vocabulary capacity rather than on the representation itself.

\begin{table}[t]
\centering\small
\begin{tabular}{@{}lrrrr@{}}
\toprule
\textbf{Scheme} & \textbf{NER} & \textbf{Sent.} & \textbf{News} & \textbf{POS}\\
\midrule
\scheme{baseline}                 & 61.10 & 54.87 & 90.98 & 94.26\\\\
\scheme{splinter\_only}           & 62.98 & 52.76 & 92.09 & 94.20\\
\scheme{syll\_then\_spl}      & \textbf{63.60} & 49.04 & \textbf{92.31} & 94.00\\
\scheme{cccvvv\_then\_spl}        & 61.51 & \textbf{56.43} & 92.03 & \textbf{94.28}\\
\scheme{spl\_then\_syll}      & 55.40 & 41.01 & 89.91 & 91.60\\
\scheme{syllabic}                 & 53.16 & 60.70 & 92.17 & 94.22\\
\bottomrule
\end{tabular}
\caption{Amharic, large model, 64k vocabulary. NER is entity-F1; other tasks weighted-F1.}
\label{tab:large64}
\end{table}

\subsection{Amharic}

Tables~\ref{tab:large64} and~\ref{tab:large16} report the large model; the small-model
results appear in \autoref{app:small}.
Across tasks, \scheme{splinter\_then\_syl}, which reverses
decomposition and reduction, is the weakest condition on every task, confirming that
decomposition must precede reduction.

Within tasks we encounter nuanced findings.
On \textbf{news classification}, every reduction scheme exceeds the
baseline at both vocabulary sizes, by one to two points.
This is the most stable effect we observe.
For \textbf{named-entity recognition}, both vocabulary sizes present advantage for \Sys{} but the magnitude and the leading scheme differ.
At 16k, \scheme{splinter\_only} reaches 65.80 against a baseline of 62.79; while at 64k,
\scheme{syllabic\_then\_spl} reaches 63.60 against 61.10.
In \textbf{sentiment analysis}, \scheme{syllabic\_then\_spl} falls below baseline at both
sizes, while \scheme{cccvvv\_then\_spl} and \scheme{syllabic} exceed it at 64k.
This is the least stable task in our experiments and we draw no ordering from it.
The \textbf{part-of-speech tagging} results, when excluding the reversed-order ablation, all fall in the 93.44--94.69 range.
At this level, the remaining differences are not negligible in relative terms, as the range represents a delta of roughly 20\% in error rate.
The ordering nevertheless varies across configurations, and the gaps fall within the spread observed across runs, so we do not rank the schemes on this task.
Finally, \autoref{tab:qa} shows that model performance on \textbf{question answering} is preserved rather than harmed: \scheme{syllabic\_then\_spl} attains the highest F1, and \scheme{splinter\_only} the highest exact-match score.

\begin{table}[t]
\centering\small
\begin{tabular}{@{}lrrrr@{}}
\toprule
\textbf{Scheme} & \textbf{NER} & \textbf{Sent.} & \textbf{News} & \textbf{POS}\\
\midrule
\scheme{baseline}                 & 62.79 & \textbf{57.02} & 89.91 & 94.09\\\\
\scheme{splinter\_only}           & \textbf{65.80} & 55.31 & \textbf{91.91} & \textbf{94.69}\\
\scheme{syll\_then\_spl}      & 63.18 & 52.14 & 91.75 & 93.98\\
\scheme{cccvvv\_then\_spl}        & 60.20 & 50.30 & 88.82 & 94.43\\
\scheme{spl\_then\_syll}      & 60.15 & 43.63 & 86.91 & 90.83\\
\scheme{syllabic}                 & 60.23 & 46.15 & 87.69 & 93.44\\
\bottomrule
\end{tabular}
\caption{Amharic, large model, 16k vocabulary.}
\label{tab:large16}
\end{table}

\begin{table}[t]
\centering\small
\begin{tabular}{@{}lrr@{}}
\toprule
\textbf{Scheme} & \textbf{EM} & \textbf{F1}\\
\midrule
\scheme{baseline}            & 47.49 & 60.55\\
\scheme{splinter\_only}      & \textbf{48.50} & 60.00\\
\scheme{syllabic\_then\_spl} & 45.00 & \textbf{61.20}\\
\bottomrule
\end{tabular}
\caption{Extractive QA on AmQA. Answers are scored in raw Amharic after inverse
transformation, which reversibility makes possible.}
\label{tab:qa}
\end{table}

\subsection{Tigrinya and Ge'ez}

\begin{table}[t]
\centering \small
\setlength{\tabcolsep}{3pt}
\begin{tabular}{@{}lrrrr@{}}
\toprule
 & \multicolumn{3}{c}{\textbf{Tigrinya}} & \textbf{Ge'ez} \\
\textbf{Scheme} & \textbf{Sent.} & \textbf{News} & \textbf{POS} & \textbf{POS}\\
\midrule
\multicolumn{5}{@{}l}{\emph{64k vocabulary}}\\
\schemefns{baseline}                  & \textbf{86.27} & 85.31 & 96.70 & 36.59\\
\schemefns{splinter\_only}            & 85.85 & 83.89 & 96.12 & 37.94\\
\schemefns{syll\_then\_spl} & 86.13 & \textbf{86.12} & 96.21 & 37.89\\
\schemefns{cccvvv\_then\_spl}    & 85.32 & 84.40 & \textbf{96.82} & \textbf{38.34}\\
\midrule
\multicolumn{5}{@{}l}{\emph{16k vocabulary}}\\
\schemefns{baseline} & 86.81 & 82.10 & \textbf{96.62} & 37.00\\
\schemefns{splinter\_only}  & \textbf{86.92} & 81.30 & 96.46 & \textbf{38.56}\\
\schemefns{syll\_then\_spl} & 86.46 & \textbf{82.67} & 96.44 & 36.38\\
\schemefns{cccvvv\_then\_spl} & 85.44 & 81.47 & 96.61 & 36.95\\
\bottomrule
\end{tabular}
\caption{Tigrinya and Ge'ez, large model. Ge'ez is transferred zero-shot from Amharic.} 
\label{tab:multiling}
\end{table}

\autoref{tab:multiling} reports the large model on Tigrinya and on Ge'ez part-of-speech tagging, the latter transferred zero-shot from Amharic.

The pattern differs from Amharic.
On Tigrinya, transformed representations achieve top performance at 64k vocabulary, with \scheme{syllabic\_then\_splinter} leading on news classification and \scheme{cccvvv\_then\_splinter} leading on part-of-speech tagging. 
At 16k vocabulary, \scheme{splinter\_only} leads on sentiment while \scheme{syllabic\_then\_splinter} leads on news classification.
Overall differences across schemes remain under two points across all Tigrinya tasks.
Ge'ez part-of-speech tagging sits between 36 and 39 for all conditions, which reflects the difficulty of zero-shot transfer rather than any property of the schemes. 
We therefore restrict our core claims to Amharic.

\section{Analysis}
\label{sec:analysis}

\paragraph{Where reduction helps.} Reduction improves news classification  and named-entity recognition at both vocabulary
sizes, while sentiment is mixed. A plausible reading is
that entity and topic labels depend chiefly on stem identity, which reduction consolidates by
mapping inflected variants onto shared forms, whereas sentiment depends more on affixal cues
such as negation, which reduction compresses into opaque markers.

\paragraph{Arrangement and ordering.} The two \scheme{X\_then\_spl} schemes differ only in
symbol layout, and they differ downstream: the interleaved form leads on named-entity
recognition at 64k, while the grouped form leads on sentiment. 
Intrinsically, the interleaved form is more token-efficient and the grouped form scores higher on R\'enyi efficiency, so the two arrangements trade off against each other.
Operation order, by contrast, is not a trade-off: reversing decomposition and reduction degrades every task, since a position-indexed reduction map cannot be applied to a stream whose structure has already changed.

\paragraph{Intrinsic measures disagree.} Tokens per word and R\'enyi efficiency select
different schemes, and neither ordering matches the downstream ordering. 
\citet{teklehaymanot-etal-2025-movoc} report a comparable dissociation, with morphology-aware segmentation improving intrinsic measures without yielding significant translation gains.
We therefore treat intrinsic statistics as descriptions of the representation rather than as predictors of task performance.

\section{Conclusion}
\label{sec:conclusion}

In this work, we introduced \Sys{}, a pre-tokenization method for subword tokenizers that decomposes Ethiopic abugida characters into explicit consonant–vowel representations, exposing linguistic structure that is otherwise fused within a single symbol.
Building on this representation, we adapted the \splinter{} reduction algorithm from abjads to an abugida and conducted a controlled study of tokenization for Ethiopic languages, holding the corpus, architecture, training budget, and evaluation protocol fixed so that tokenization remained the sole variable.

Four findings emerge.
First, exposing the consonantal skeleton helps tasks that turn on stem identity, consistently for news classification and for named-entity recognition, while tasks that depend on the inflectional material reduction compresses benefit less or not at all; the effect of pre-tokenization is therefore task-dependent rather than uniform.
Second, how the two operations are composed matters as much as whether they are applied, since reversing decomposition and reduction degrades every task.
Third, different vocabulary sizes provide different preferences, so intended use case is a factor in deciding on a pre-tokenization schema.
Fourth, the intrinsic measures commonly used to select tokenizers disagree with one another here and do not order the schemes as downstream accuracy does, so they are better read as descriptions of a representation than as predictions of its usefulness.

Future work might extend evaluation for Tigrinya and Ge'ez beyond the scope reported here, investigate additional languages using the Ethiopic script, and evaluate this approach on other abugida writing systems. 
Additionally, introducing a curated affix inventory could allow corpus frequency to validate rather than define reductions, and testing generative decoding will confirm if reversibility is preserved when a model generates sequences unseen during encoding.

\section*{Limitations}

 Language coverage is uneven. 
Amharic is evaluated on four tasks; Tigrinya and Ge'ez are not fully covered in the tables reported here, which reflects dataset availability rather than design.
Claims about Ethiopic languages therefore more takes on Amharic evidence.
The interpretation offered in Section~\ref{sec:analysis}, that reduction consolidates stems at the expense of affixes, is consistent with the observed pattern but was not verified against a morphological analyser.
Reductions are learned from corpus frequency and have no access to morphological knowledge, so a frequent reduction need not correspond to a morpheme.
Finally, our results concern more on a single corpus and the family of subword algorithms.
Whether the same trade appears with different pretraining data, or under character or byte-level models that do not tokenize at all, is untested.

\section*{Ethics Statement}
This work concerns tokenization for Amharic, Tigrinya and Ge'ez, languages with limited NLP
resources relative to their speaker populations, and improved tokenization may lower the cost of building language technology for these communities. 
All data used is in these three languages, written in the Ge'ez script, and is drawn from news articles, social media posts and Wikipedia; annotation was performed by native speakers as reported by the original dataset creators.
Two cautions apply.
Evaluation coverage is uneven across the three languages, so benefits should not be assumed to transfer equally. 
And reductions are learned from corpus statistics, so a rule frequent in news text need not be linguistically general.

\section*{Acknowledgments}
This research was supported by the Israel Science Foundation (grant No. 1166/23).

\bibliography{custom}

\newpage

\appendix
\section{Decomposition of the \emph{s-b-r} Family}
\label{app:sbr}

Examples of consonant--vowel decomposition applied to the \emph{s-b-r} morphological family are presented in \autoref{tab:sbr}.

\begin{table*}[t]
\centering\small
\begin{tabular}{@{}l l l l@{}}
\toprule
\textbf{Form} & \textbf{Gloss} & \textbf{Consonants} & \textbf{Decomposed}\\
\midrule
\gz{sbr_broke}   & \say{he broke}       & \gz{se}\gz{be}\gz{re}
  & \gz{se} \gz{be} \gz{re}\\
\gz{sbr_frag}    & \say{fragments}      & \gz{se}\gz{be}\gz{re}\gz{be}\gz{re}
  & \gz{se}$V_5$ \gz{be}$V_5$ \gz{re}$V_5$ \gz{be}$V_3$ \gz{re}$V_2$\\
\gz{sbr_shatter} & \say{he shattered}   & \gz{se}\gz{be}\gz{be}\gz{re}
  & \gz{se} \gz{be}$V_3$ \gz{be} \gz{re}\\
  \gz{sbrpl} & \say{they shattered}   & \gz{se}\gz{be}\gz{be}\gz{re}
  & \gz{se} \gz{be}$V_3$ \gz{be} \gz{re}$V_1$\\
\gz{sbr_passive} & \say{it was broken}  & \gz{te}\gz{se}\gz{be}\gz{re}
  & \gz{te} \gz{se} \gz{be} \gz{re}\\
\gz{sbr_instr}   & \say{instrument for breaking} & \gz{me}\gz{se}\gz{be}\gz{re}\gz{ye}
  & \gz{me} \gz{se} \gz{be} \gz{re}$V_2$ \gz{ye}$V_3$\\
\bottomrule
\end{tabular}
\caption{The \emph{s-b-r} family after consonant--vowel decomposition. Each row gives a derived form, its gloss, the base consonants it decomposes to, and the decomposed sequence, in which every consonant appears in canonical first order followed by a marker for any non-first vowel order. 
In the surface fid"al of the first column, the six words share almost no characters, since each template realises the same consonants in different vowel orders. 
After decomposition, the root consonants \gz{se}--\gz{be}--\gz{re} become explicit in every form, either directly or interleaved with vowel markers and additional morphological material. Reduction can then iteratively remove this surrounding material—the passive prefix \gz{te}, the instrumental \gz{me} with its suffix \gz{ye}, and the reduplicated \gz{be} of the frequentative—and, where the resulting form is attested, reach the root \gz{sbr_broke}. Forms whose consonants are all first order carry no marker at all.}
\label{tab:sbr}
\end{table*}

\section{All Pre-tokenization Pipelines}
\label{app:pipelines}

\begin{figure*}[!ht]
\centering
\small
\begin{tikzpicture}[
  scale=0.66, transform shape,
  st/.style={draw, rounded corners=1.5pt, minimum height=6.5mm, inner xsep=4pt,
             fill=gray!10, font=\small},
  tk/.style={draw, minimum height=6mm, minimum width=12mm, font=\small},
  nm/.style={font=\small\ttfamily, anchor=east},
  note/.style={font=\scriptsize\itshape, text=black!65, anchor=north},
  ar/.style={-{Latex[length=1.7mm]}, thick}]
\def\R{-1.5}

\node[nm] at (0,0) {baseline};
\node[st] (t0) at (2.0,0) {\gz{sbpl}};
\node[tk] (k0) at (15.6,0) {tokenizer};
\draw[ar] (t0) -- (k0);
\node[note] at (2.0,-0.45) {original text};

\node[nm] at (0,\R) {\splinter{}\_only};
\node[st] (t1) at (2.0,\R) {\gz{sbpl}};
\node[st] (s1) at (4.5,\R) {\splinter{}};
\node[st] (r1) at (7.6,\R) {\gz{sbru}\,$R$};
\node[tk] (k1) at (15.6,\R) {tokenizer};
\draw[ar] (t1) -- (s1); \draw[ar] (s1) -- (r1); \draw[ar] (r1) -- (k1);
\node[note] at (7.6,\R-0.45) {reduced, but not to the root};

\node[nm] at (0,2*\R) {syllabic};
\node[st] (t2) at (2.0,2*\R) {\gz{sbpl}};
\node[st] (s2) at (4.5,2*\R) {\Sys{}};
\node[st] (r2) at (7.6,2*\R) {\gz{se}\,\gz{be}$V_3$\,\gz{be}\,\gz{re}$V_1$};
\node[tk] (k2) at (15.6,2*\R) {tokenizer};
\draw[ar] (t2) -- (s2); \draw[ar] (s2) -- (r2); \draw[ar] (r2) -- (k2);
\node[note] at (7.6,2*\R-0.45) {decomposed, no reduction};

\node[nm] at (0,3*\R) {syllabic\_then\_\splinter{}};
\node[st] (t3) at (2.0,3*\R) {\gz{sbpl}};
\node[st] (a3) at (3.9,3*\R) {\Sys{}};
\node[st] (m3) at (6.4,3*\R) {\gz{se}\,\gz{be}$V_3$\,\gz{be}\,\gz{re}$V_1$};
\node[st] (b3) at (9.0,3*\R) {\splinter{}};
\node[st] (r3) at (11.6,3*\R) {\gz{sberu}\,$R_1R_2$};
\node[tk] (k3) at (15.6,3*\R) {tokenizer};
\draw[ar] (t3) -- (a3); \draw[ar] (a3) -- (m3); \draw[ar] (m3) -- (b3);
\draw[ar] (b3) -- (r3); \draw[ar] (r3) -- (k3);
\node[note] at (11.6,3*\R-0.45) {root recovered; reduction markers carry positions};

\node[nm] at (0,4*\R) {cccvvv\_then\_\splinter{}};
\node[st] (t4) at (2.0,4*\R) {\gz{sbpl}};
\node[st] (a4) at (4.3,4*\R) {\Sys{} (grouped)};
\node[st] (m4) at (7.0,4*\R) {\gz{sbdecg}\;$V_3V_1$};
\node[st] (b4) at (9.4,4*\R) {\splinter{}};
\node[st] (r4) at (12.0,4*\R) {\gz{sberu}\,$R_1R_2$};
\node[tk] (k4) at (15.6,4*\R) {tokenizer};
\draw[ar] (t4) -- (a4); \draw[ar] (a4) -- (m4); \draw[ar] (m4) -- (b4);
\draw[ar] (b4) -- (r4); \draw[ar] (r4) -- (k4);
\node[note] at (12.0,4*\R-0.45) {grouped, reduced by position};

\node[nm] at (0,5*\R) {\splinter{}\_then\_syllabic};
\node[st] (t5) at (2.0,5*\R) {\gz{sbpl}};
\node[st] (a5) at (4.3,5*\R) {\splinter{}};
\node[st] (m5) at (7.0,5*\R) {\gz{sbru}\,$R$};
\node[st] (b5) at (9.4,5*\R) {\Sys{}};
\node[st] (r5) at (11.8,5*\R) {\gz{sberu}$V_1$\,$R$};
\node[tk] (k5) at (15.6,5*\R) {tokenizer};
\draw[ar] (t5) -- (a5); \draw[ar] (a5) -- (m5); \draw[ar] (m5) -- (b5);
\draw[ar] (b5) -- (r5); \draw[ar] (r5) -- (k5);
\node[note] at (11.8,5*\R-0.45) {decomposition of the already-reduced form};

\node[tk, minimum width=10mm] (lm) at (17.8,2.5*\R) {LM};
\foreach \k in {k0,k1,k2,k3,k4,k5} { \draw[ar] (\k) -- (lm); }
\node[note, anchor=south] at (16.8,2.5*\R+0.45) {embeddings};
\end{tikzpicture}
\caption{The six pre-tokenization conditions applied to \gz{sbpl}
(\emph{s\"ababb\"aru}, \say{they shattered}).
$V\!n$ marks vowel order $n$.
Reduction markers are written $R_1$, $R_2$ for legibility; each is a single codepoint in the
implementation, recording what was deleted and where.
Here $R_1$ records the removal of the reduplicated \gz{be} together with its vowel order $V_3$,
and $R_2$ that the final \gz{re} was lowered from $V_1$ to first order, while $R$ in
\SplOnly{} records the removal of the syllable \gz{ba}.
\SplOnly{} reduces the surface form but cannot expose the root, since on an abugida the
vowels remain fused inside the fid\"al characters; only after decomposition are the root
consonants \gz{se}\gz{be}\gz{re} visible as contiguous symbols.
Decomposition exposes structure and reduction exploits it, which is why the two are separate
contributions, and why \SplSyl{} --- which reduces first --- is included as an ablation.
}
\label{fig:allpipelines}
\end{figure*}
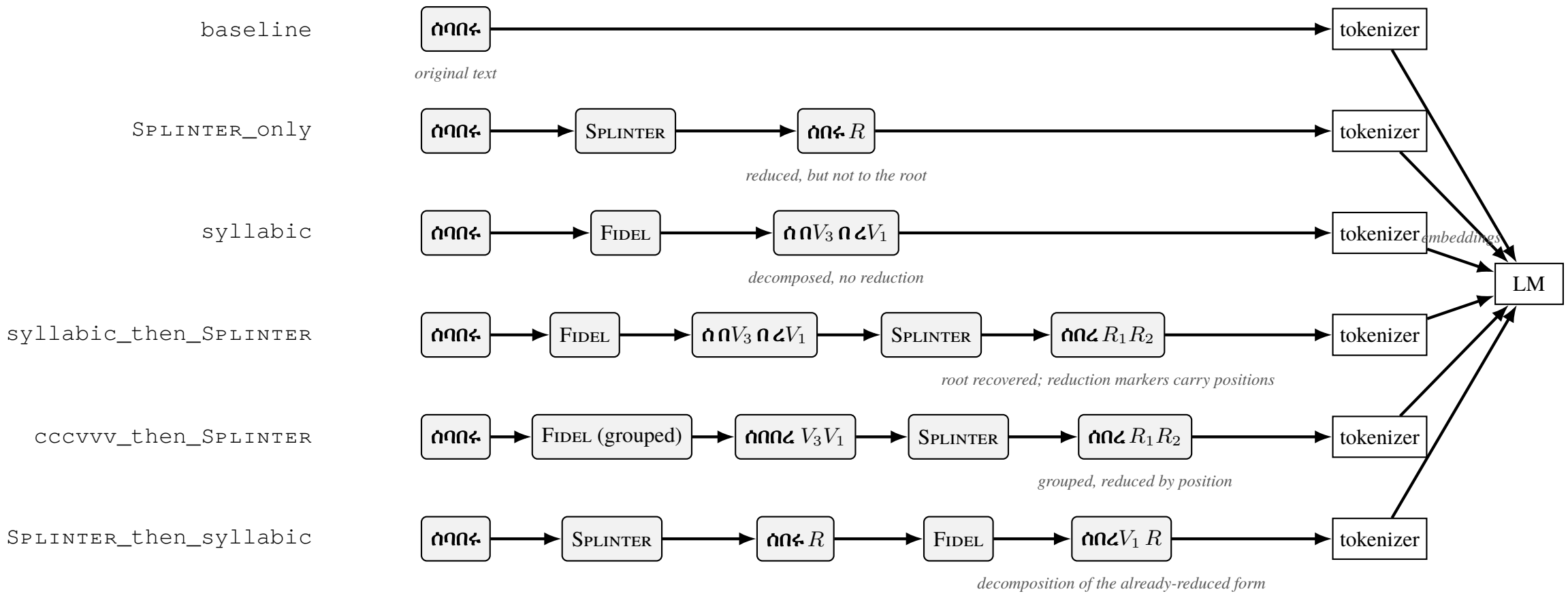

The six conditions are shown as pipelines in \autoref{fig:allpipelines}.

\section{Decoding Procedure}
\label{app:decoding}

\begin{definition}[CV manifold]
Let $\mathcal{M}$ be the set of all valid Ge'ez strings that can be written as sequences of
pairs $(c_i, v_i)$ with $c_i \in C$ a base consonant and $v_i$ a valid vowel order.
\end{definition}

Given a token sequence $S$ produced by the model, decoding finds the nearest valid word:
\begin{equation*}
\hat{W} = \arg\min_{W \in \mathcal{M}} d(S, W).
\end{equation*}
Let $C = [c_1,\dots,c_n]$ be the consonants extracted from $S$ and
$V = \{(p_j, m_j)\}_{j=1}^{m}$ the vowel instructions decoded from marker tokens, with $p_j$
a position and $m_j$ a marker. The distance is
\begin{equation}
\begin{split}
 d(S,W) = \min_{\alpha \in A} \textstyle\sum_{j=1}^{m} |\alpha(j)-p_j| \\
 {} + \lambda\max(0,m-n) + \mu\max(0,n-m),
\end{split}
\end{equation}
where $A$ is the set of monotone, injective assignments from instructions to positions. The
first term aligns vowel instructions to consonant positions, while the second and third
penalise overflow and underflow. Letting $\lambda$ and $\mu$ tend to infinity enforces strict
handling: surplus instructions are discarded and unfilled positions take the default order.
All schemes are validated by round-trip reconstruction on held-out Amharic and reproduce the
original text exactly.

\section{UnigramLM Intrinsic Statistics}
\label{app:intrinsic}

Intrinsic benchmark results for UnigramLM across vocabulary sizes are presented in \autoref{tab:unigram}.
\begin{table*}[t]
\centering\small
\begin{tabular}{@{}lrrrrr@{}}
\toprule
\textbf{Scheme} & \textbf{\begin{tabular}[c]{@{}r@{}}R\'enyi\\ efficiency\end{tabular}} & \textbf{\begin{tabular}[c]{@{}r@{}}Tokens\\ per word\end{tabular}} & \textbf{\begin{tabular}[c]{@{}r@{}}4+ token\\ words\end{tabular}} & \textbf{\begin{tabular}[c]{@{}r@{}}1-char\\ tokens\end{tabular}} & \textbf{\begin{tabular}[c]{@{}r@{}}Distinct\\ Neighbors\end{tabular}} \\
\midrule
\multicolumn{6}{@{}l}{\emph{64k vocabulary}}\\
\scheme{baseline}                  & 0.547 & 1.382 & 2.67\% & 2.53\% & 6988\\
\scheme{syllabic\_then\_\splinter{}} & 0.574 & \textbf{1.346} & \textbf{2.49\%} & \textbf{1.32\%} & 5865\\
\scheme{\splinter{}\_then\_syllabic} & \textbf{0.600} & 1.377 & 2.59\% & 1.43\% & 7104\\
\scheme{\splinter{}\_only}            & 0.598 & 1.382 & 2.81\% & 2.43\% & 7082\\
\scheme{cccvvv\_then\_\splinter{}}    & 0.596 & 1.503 & 6.17\% & 3.14\% & 7572\\
\scheme{syllabic}                  & 0.548 & 2.422 & 5.36\% & 1.69\% & \textbf{5766}\\
\midrule
\multicolumn{6}{@{}l}{\emph{32k vocabulary}}\\
\scheme{baseline}                  & 0.561 & 1.525 & \textbf{4.59\%} & 3.90\% & 6370\\
\scheme{syllabic\_then\_\splinter{}} & 0.602 & \textbf{1.482} & 4.70\% & \textbf{1.81\%} & 5636\\
\scheme{\splinter{}\_then\_syllabic} & 0.625 & 1.519 & 4.67\% & 1.83\% & 6348\\
\scheme{\splinter{}\_only}            & 0.623 & 1.529 & 5.09\% & 3.91\% & 6354\\
\scheme{cccvvv\_then\_\splinter{}}    & \textbf{0.633} & 1.707 & 10.90\% & 4.12\% & 6473\\
\scheme{syllabic}                  & 0.541 & 2.556 & 8.64\% & 2.10\% & \textbf{4818}\\
\midrule
\multicolumn{6}{@{}l}{\emph{16k vocabulary}}\\
\scheme{baseline}                  & 0.574 & 1.730 & \textbf{7.87\%} & 5.17\% & 4990\\
\scheme{syllabic\_then\_\splinter{}} & 0.638 & \textbf{1.697} & 8.57\% & 2.65\% & 4613\\
\scheme{\splinter{}\_then\_syllabic} & 0.654 & 1.731 & 8.45\% & \textbf{2.56\%} & 4861\\
\scheme{\splinter{}\_only}            & 0.651 & 1.747 & 9.07\% & 5.41\% & 4845\\
\scheme{cccvvv\_then\_\splinter{}}    & \textbf{0.672} & 1.991 & 17.22\% & 5.48\% & 4948\\
\scheme{syllabic}                  & 0.530 & 2.705 & 13.28\% & 2.72\% & \textbf{3654}\\
\midrule
\multicolumn{6}{@{}l}{\emph{8k vocabulary}}\\
\scheme{baseline}                  & 0.583 & \textbf{2.039} & \textbf{14.12\%} & 7.38\% & 3335\\
\scheme{\splinter{}\_then\_syllabic} & 0.674 & 2.086 & 15.90\% & \textbf{4.24\%} & 3043\\
\scheme{syllabic\_then\_\splinter{}} & 0.673 & 2.081 & 16.60\% & 4.52\% & 2977\\
\scheme{\splinter{}\_only}            & 0.674 & 2.118 & 17.08\% & 8.43\% & 2929\\
\scheme{cccvvv\_then\_\splinter{}}    & \textbf{0.701} & 2.410 & 26.34\% & 7.72\% & 3368\\
\scheme{syllabic}                  & 0.513 & 2.902 & 20.46\% & 3.98\% & \textbf{2407}\\
\bottomrule
\end{tabular}
\caption{Intrinsic statistics for UnigramLM across vocabulary sizes. Bold marks the best
value per column within each block. The ordering reproduces the BPE results of
Table~\ref{tab:intrinsic}.}
\label{tab:unigram}
\end{table*}
\section{Small-Model Results}
\label{app:small}

Downstream evaluation results for the Amharic small model across tasks and vocabulary sizes are presented in \autoref{tab:small}.

\begin{table*}[t]
\centering\small
\begin{tabular}{@{}lrrrr@{}}
\toprule
\textbf{Scheme} & \textbf{NER} & \textbf{Sent.} & \textbf{News} & \textbf{POS}\\
\midrule
\multicolumn{5}{@{}l}{\emph{16k vocabulary}}\\
\scheme{baseline}                  & \textbf{33.59} & \textbf{47.50} & 88.36 & 92.31\\
\scheme{\splinter{}\_only}           & 32.91 & 47.00 & \textbf{89.93} & 89.86\\
\scheme{syllabic\_then\_\splinter{}} & 32.21 & 41.03 & 89.68 & 89.81\\
\scheme{cccvvv\_then\_\splinter{}}   & 25.99 & 40.97 & 87.87 & 90.74\\
\scheme{\splinter{}\_then\_syllabic} & 27.56 & 28.65 & 82.05 & 87.62\\
\scheme{syllabic}                  & 21.62 & 47.11 & 88.35 & \textbf{92.63}\\
\midrule
\multicolumn{5}{@{}l}{\emph{64k vocabulary}}\\
\scheme{baseline}                  & \textbf{36.43} & 49.41 & 91.49 & 90.73\\
\scheme{\splinter{}\_only}           & 35.12 & 46.97 & \textbf{92.04} & \textbf{92.28}\\
\scheme{syllabic\_then\_\splinter{}} & 34.37 & 47.27 & 91.25 & 90.55\\
\scheme{cccvvv\_then\_\splinter{}}   & 32.86 & \textbf{50.18} & 89.31 & 90.43\\
\scheme{\splinter{}\_then\_syllabic} & 27.03 & 47.20 & 89.44 & 89.78\\
\scheme{syllabic}                  & 19.73 & 46.60 & 90.31 & 90.32\\
\bottomrule
\end{tabular}
\caption{Amharic, small model. At this capacity the baseline is competitive on named-entity recognition, while the reduction schemes retain their advantage on news classification.}
\label{tab:small}
\end{table*}

\section{Training Hyperparameters}
\label{app:training-hyperparameters}

\paragraph{Pretraining.}
All encoders are pretrained from scratch using masked language
modeling. We use AdamW with linear warmup over the first 6\% of
steps followed by cosine learning-rate decay. The small models use
a learning rate of $1\times10^{-4}$ and a
batch size of 16, while the large models use a learning rate of
$5\times10^{-5}$ and a batch size of 32. The maximum training
duration is 25 epochs for small models and 30 epochs for large
models. The sequence length is 512 tokens.

\paragraph{Fine-tuning.}
For downstream tasks, we fine-tune the full encoder with a learning
rate of $2\times10^{-5}$ for 5 epochs, except for question answering,
which uses 10 epochs. We select the checkpoint with the best
validation performance. All experiments use three random seeds.

\end{document}